\documentclass[letterpaper, 10 pt, conference]{ieeeconf}  

\IEEEoverridecommandlockouts                              
\usepackage{amsmath} 
\usepackage{amssymb}  
\usepackage{cite}
\usepackage{threeparttable}
\usepackage{stfloats}
\usepackage{comment}

\title{\LARGE \bf
Zero-shot Deep Reinforcement Learning Driving Policy Transfer for Autonomous Vehicles based on Robust Control 
}

\author{Zhuo Xu$^{*}$, Chen Tang$^{*}$ and Masayoshi Tomizuka
\thanks{{$^\dagger$}This work was supported in part by Berkeley DeepDrive (BDD).}
\thanks{{$^\dagger$}Z. Xu, C. Tang and M. Tomizuka are with the Department of Mechanical
Engineering, University of California, Berkeley, CA 94720 USA (e-mail:
zhuoxu@berkeley.edu, chen\_tang@berkeley.edu, tomizuka@berkeley.edu).}
\thanks{*These authors contributed equally to this work.}
}

\usepackage{algorithm,algpseudocode}
\usepackage{graphicx}\usepackage{multirow}
\usepackage{url}
\begin{document}

\maketitle
\thispagestyle{empty}
\pagestyle{empty}

\begin{abstract}
Although deep reinforcement learning (deep RL) methods have lots of strengths that are favorable if applied to autonomous driving, real deep RL applications in autonomous driving have been slowed down by the modeling gap between the source (training) domain and the target (deployment) domain. Unlike current policy transfer approaches, which generally limit to the usage of uninterpretable neural network representations as the transferred features, we propose to transfer concrete kinematic quantities in autonomous driving. The proposed robust-control-based (RC) generic transfer architecture, which we call RL-RC, incorporates a transferable hierarchical RL trajectory planner and a robust tracking controller based on disturbance observer (DOB). The deep RL policies trained with known nominal dynamics model are transfered directly to the target domain, DOB-based robust tracking control is applied to tackle the modeling gap including the vehicle dynamics errors and the external disturbances such as side forces. We provide simulations validating the capability of the proposed method to achieve zero-shot transfer across multiple driving scenarios such as lane keeping, lane changing and obstacle avoidance.
\end{abstract}

\section{Introduction}
Learning intelligent and reliable driving policies has been an ongoing challenge for both deep learning and control. Although conventional planning-control \cite{levinson2011towards} and imitation-oriented learning \cite{nvidia2016end}\cite{xu2016end} approaches are shown to have capability of controlling autonomous vehicles, deep reinforcement-learning-based (deep RL) methods are promising to tackle more complicated and interacting scenarios that conventional methods are incapable to solve, as well as rare cases outside the demonstration dataset for the supervised imitation learning.

While the vast exploration and flexible hierarchical configurations enable deep RL methods to obtain versatile policies more easily \cite{rusu2016progressive}\cite{xu2017cascade}, robustness has essentially been the main drawback that prevents the application of deep RL in autonomous driving. Concretely, pre-optimized deep RL policies are over specialized, and thus often fail when the target vehicle has dynamics variation from the training setting, or when the target vehicle is affected by force disturbances due to strong wind or body incline. These differences between the source and target settings together are called the modeling gap. Given the modeling gap being an inevitable barrier for the deployment of deep RL in autonomous driving, in this contribution we aim to bridge the modeling gap by achieving fast and safe transfer of deep RL autonomous driving policies.

Prior efforts in deep RL attempt to achieve such objective using transfer learning and meta learning. Various contributions in these areas have indeed brought source policies to work in the target settings, but their applications in autonomous driving are limited due to safety concerns. Overall, such deep RL transfer methods embed transferable representations into uninterpretable neural networks and hope for the best in the target domain, and thus are not transparent and reliable. We seek to solve the transfer problem using an alternative tool: robust control (RC), and propose a generic RL-RC transfer framework. In this framework, the deep RL policy is applied to an imaginary setting in the source domain to generate a reference trajectory for the target vehicle. A robust controller is applied to track the reference trajectory tolerating the modeling gap. This framework is generic in that it can be generalized to any deep RL control policy transfer tasks, and that various kinds of robust controllers can be used for tracking. Compared to other transfer learning methods, the proposed framework has two fundamental advantages: (i) the transfer of the interpretable kinematic features makes the transfer framework transparent and reliable; (ii) stability and response in time and frequency domains, can be well defined and analyzed.

The contribution in this paper is four-fold. First, we purpose a generic approach for deep RL driving policy transfer. Second, we implement a hierarchical RL model which serves as the transferable trajectory planner. Third, we develop a disturbance-observer-based (DOB) robust tracking controller so as to actively reject the disturbances induced by the modeling gap. Finally, we report simulation results validating that the RL-RC architecture can zero-shoot transfer the policy under certain level of parameter variation and external disturbances.

The rest of this paper is organized as follows: The related work is summarized in Section II. In Section III, the generic RL-RC architecture for policy transfer is described. Then, the setting of our autonomous driving tasks, together with the implementation of all the components in the RL-RC architecture, is illustrated. In Section IV, simulation results are presented and evaluated to show the effectiveness of the proposed method, followed by the conclusion in Section V.

\section{Related Work}
{\bf Deep Learning for Autonomous Driving.} Prior works on deep-learning-based autonomous driving policy have mostly been developed based on behavior cloning or imitation learning. From the pioneering ALVINN \cite{pomerleau1989alvinn} to recent works such as \cite{nvidia2016end}, \cite{xu2016end} and \cite{muller2006off}, researchers trained neural networks to predict driving actions. However, the models were trained in a supervised manner, and are inevitably limited by the dataset. Deep RL-based driving policy learning methods have demonstrated their capability in simulators like TORCS. Various deep RL methods including deep Q-learning and actor-critic algorithms have been applied to train lane-keeping policies with low-dimensional feature vector or image pixels as inputs \cite{lillicrap2015continuous,mnih2016asynchronous,sallab2016end}. Meanwhile, designing  transfer methods to bridge the modeling gao for deep RL application in autonomous driving is a open but intriguing problem.

{\bf Deep RL Policy Transfer.} One category of transfer learning uses source domain randomization to embed robustness. The EPOpt algorithm \cite{rajeswaran2016epopt} randomly samples dynamic parameters from a prior distribution and optimizes the policy for cases with worst-performing dynamic parameters. \cite{peng2017sim} uses RNN to take in historical paths, expecting the policy to be adaptive by implicitly identifying the parameters. Similar idea has been adopted in \cite{yu2017preparing}, but an on-line system identification module is explicitly introduced to identify the parameters. The estimated parameters are fed into a parameter-universal policy. All these works are limited in that they only handle the discrepancy of dynamic parameters, while robustness against external disturbances is excluded from consideration.

Another line of work is through model adaptation. \cite{fu2016one} trains a neural network for dynamics model and adapts its local linear model on-line for model-based control methods. In \cite{christiano2016transfer}, a deep inverse dynamics model of the target system is trained, and transfer is achieved by executing inverse model outputs to reach the nominal state generated by performing source policy in source environment. One problem of such approaches is that, unlike tracking by feedback control, a feasible and accurate inverse model for feedforward tracking is hardly guaranteed in the target domain. Also, a neural network approximating system dynamics is non-trivial to find in practice. Furthermore, fine-tuning in the target domain is not desirable due to safety. 

A method developed on a similar idea to the one in this paper is in \cite{harrison2017adapt}, in which a MPC controller is designed to stabilize the target system around the nominal trajectory generated by consecutively applying the policy in the source system. Theorems on tube-based MPC ensure that the states are bounded under certain modeling error. However, the bound cannot be explicitly found and no asymptotic stability can be guaranteed. Moreover, solving on-line optimization problem is computationally expensive, while robust controllers are usually easier and faster for the trajectory tracking problem of automated vehicles. 

{\bf DOB for Vehicle Lateral Control.} Vehicle lateral control for trajectory tracking has been well-established with many existing methods \cite{paden2016survey}. DOB is a robust control technique to reject disturbances with guaranteed robust stability for linear system \cite{chen2016disturbance}. Moreover, given a stabilizing nominal controller, Q-filter in DOB can be an arbitrary stable filter to make the sensitivity function shaped as desired \cite{chen2013control}. It has been applied to robust lateral trajectory tracking and its effectiveness has been verified by experiments \cite{rathgeber2014lateral}.     

\section{Methodology}
\subsection{Architecture}
The proposed RL-RC policy transfer architecture consists of a deep RL-based high-level planning module and a RC-based low-level tracking controller. The overall methodology consists of the off-line deep RL policy training and the on-line policy transfer. In the off-line source domain, we pretrain a deep RL policy mapping the perception input to control commands, which can drive the source vehicle to produce a trajectory completing the control task. 

\begin{figure}[t]
\centering
\includegraphics[scale=0.34]{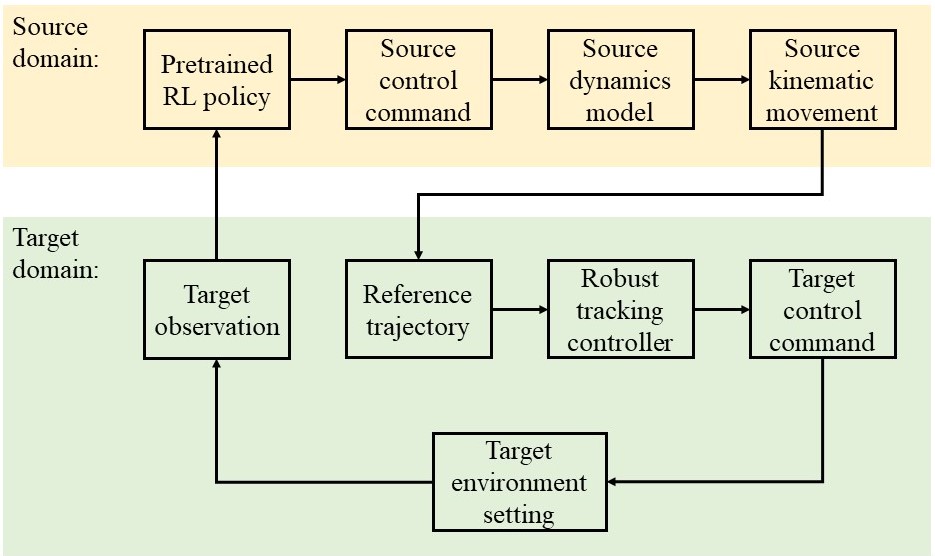}
\caption{Target domain on-line implementation of the RL-RC architecture}
\label{fig:structure}
\end{figure}

For the on-line transfer in the target domain, Fig. \ref{fig:structure} shows how the system works to transfer the source driving policies. In general, the RL-RC system performs closed-loop tracking of a finite-horizon previewed trajectory generated by the pretrained policy. Concretely, at each time step, the target agent obtains the perception observation. Then according to this observation, the system constructs an imaginary source agent in the same setting as the target agent. In the imaginary source domain, the pretrained policy is used to control the imaginary source agent to perform the driving task for a finite horizon, resulting in a trajectory of kinematic states, which serves as the reference for the target vehicle. Given the reference trajectory, the target agent uses a closed-loop robust tracking controller to produce the actual control command for the target vehicle. As the target environment makes one step forward, the observation for the new timestep is collected, and the system repeats the same procedure. In this architecture, the kinematic features are transfered without change, and the modeling gap is compensated by RC.

To generate the reference trajectory in the imaginary source domain, one can use the cascaded high-level planner, which consists of the pretrained policy and the source environment, or train a deep RL policy network that directly maps perception input to trajectory of kinematic states. We choose the latter approach so as to optimize the policy for the whole procedure, enabling both dynamical feasibility and optimality of the planned trajectory. 

There are two key underlying assumptions for the RL-RC system: (i) the trajectory planned by the imaginary source agent is comparably satisfying for the target task; (ii) it is feasible for the target vehicle to track the trajectories produced by the source vehicle. These assumptions are reasonable as the source and target vehicles and settings are similar, and the RL-RC is only effective for such cases.

Our proposed RL-RC framework is generic in many aspects: 
\begin{enumerate}
\item This system can perform policy transfer for various kinds of tasks in autonomous driving. It can also be applied in other robotics and control scenarios if the assumptions are satisfied and a robust controller can be designed. In this paper, we evaluate the RL-RC approach for three typical driving tasks: lane keeping (LC), lane changing (LC), and obstacle avoidance (OA).
\item The proposed method places no restrictions on the structures of the deep RL policy.  In this paper we use a hierarchical RL model to generate reference trajectories in the imaginary source domain.
%
%
%
%
\item Any kind of RC method is compatible with the RL-RC architecture. Among various RC algorithms, we use a disturbance-observer-based (DOB) tracking controller to actively reject the disturbance induced by modeling error.
\end{enumerate}

\subsection{Environment}
\begin{figure}[t]
\centering
\includegraphics[scale=0.40]{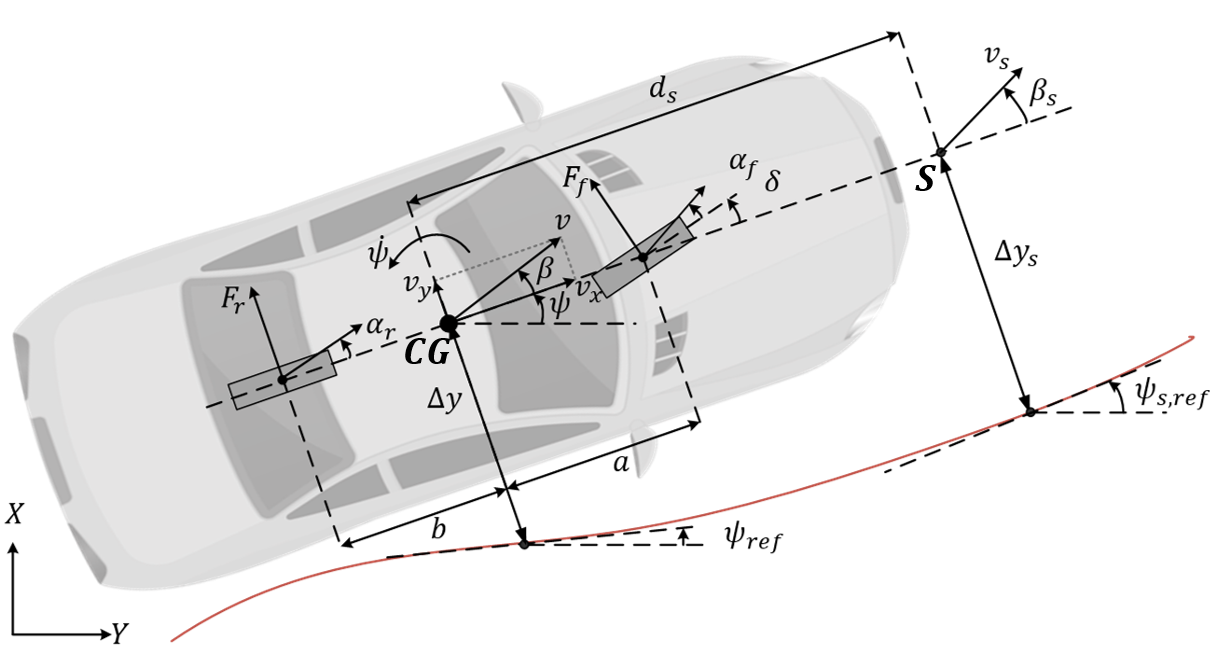}
\caption{Illustration of the terminology for the simulated environment and the linear tracking model for the controller}
\label{fig:dynamics}
\end{figure}

In both the source domain and the target domain, we simulate the vehicle dynamics using nonlinear bicycle model discretized using forward Euler method with $\Delta{t}=0.02s$. The model is illustrated in Fig. \ref{fig:dynamics}. The longitudinal acceleration $a_x$ and the derivative of steering angle $\dot\delta$ are inputs. The seven state variables are longitudinal speed $v_x$, lateral speed $v_y$, yaw rate $\omega_z$, global coordinates $X, Y$, yaw angle $\psi$ and steering angle $\delta$. To obtain simulated vehicle with high-fidelity, we include tire force saturation and restrict magnitudes of $a_x, \delta, \dot{\delta}$ to reasonable ranges. Moreover, we omit longitudinal powertrain dynamics, assuming nearly perfect longitudinal tracking by a low-level longitudinal controller.

Apart from the vehicle dynamics, we also define terms related to the driving scenarios. We define lateral displacements, $\Delta{y}$ and $\Delta{y_s}$, and yaw errors, $\Delta{\psi}$ and $\Delta{\psi_s}$, to represent the tracking errors relative to lane or reference trajectory (Fig. \ref{fig:dynamics}). $\Delta{y}$ and $\Delta{\psi}$ are defined with respect to the center of gravity (CG), and $\Delta{y_s}$ and $\Delta{\psi_s}$ are defined at a point $S$ specified by a look-ahead distance $d_s$. We denote the angle of the vehicle speed $v$ as $\psi_v=\psi+\beta$, where $\beta$ is the angle between vehicle speed and yaw. Similarly, the angle of $v_s$ is $\psi_{v_s} = \psi+\beta_s$. The yaw errors are the angle between velocity and the tangent direction of the reference curve, specifically, $\Delta\psi=\psi_v-\psi_{ref}$ and $\Delta\psi_s=\psi_{v_s}-\psi_{s,ref}$. With the yaw error $\Delta{\psi}$, one can easily derive ego vehicle's speed along and perpendicular to the reference curve: $v_{\parallel}=v\cos(\Delta{\psi})$ and $v_{\perp}=v\sin(\Delta{\psi})$.

For the collision avoidance task, we consider a simple scenario with a single surrounding vehicle, which perfectly tracks the lane with constant speed $v_{srd}$. We denote the relative longitudinal and lateral position of the surrounding vehicle by $\Delta{x_{srd}}$ and $\Delta{y_{srd}}$.

\subsection{Hierarchical RL}

So as to learn driving policy using deep RL, we first define the Markov Decision Process (MDP) for the driving tasks. Since different tasks have different observations, we define clusters of observations as entities. Concretely, we define the observation of ego vehicle to be ${o}_{vh} = [v_x\ v_y\ \omega_z\ \delta]^T$. For each lane, denoted $l_i$, $i\in N$, we define tracking observations as ${o}_{ref,i} = [\Delta{y_i}\ \Delta{\psi_i}\ \Delta{y_{s,i}}\ \Delta{\psi_{s,i}}]^T$. The lane that the ego vehicle is tracking is denoted using $l_{i^*}$. The obstacle observation is ${o}_{srd} = [v_{srd}\ \Delta{x_{srd}}\ \Delta{y_{srd}}]^T$. For simplification, we assume perfect and deterministic observations. For all the tasks, control command $a_{vh} = [a_x\ \dot{\delta}]^T$ is defined as the actions in the MDP. The actions are continuous and normalized by their maximum allowable values. 

For performance evaluation, we define the step tracking reward function at timestep $t$:
\begin{align}
r_{tra}(t) = v_{\parallel,i^*}(t)-|v_{\perp,i^*}(t)|-\eta\cdot\Delta{y_{i^*}(t)}^2,
\end{align}
where $\eta>0$ is an importance factor. The step rewards are discounted with factor $\gamma$. An episode is defined to have maximum of 1000 timesteps, The episode will be terminated by large deviation from the lane or violating collision constraint. Meanwhile, the agent will be punished with a large negative reward, denoted $r_{dev}$ and $r_{col}$. The interfaces of lane keeping (LK), lane changing (LC) and obstacle avoidance (OA) are shown in Table. \ref{driving_env_table}.

\begin{figure}[t]
\centering
\includegraphics[scale=0.38]{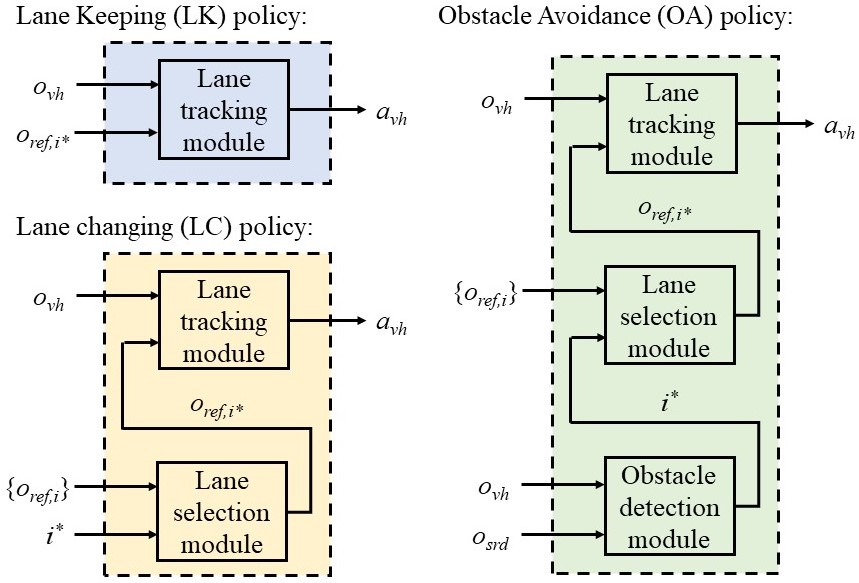}
\caption{Usage of hierarchical RL modules to  assemble policies for the lane keeping (LK), lane changing (LC) and obstacle avoidance (OA) tasks.}
\label{fig:hierarchical}
\end{figure}

In our implementation, a hierarchical RL model that can modularize distinct driving attributes is applied. The attributes refer to driving behaviors such as obstacle detection, lane selection, and lane tracking. Table \ref{rl_module_table} and Fig. \ref{fig:hierarchical} give the detailed module interfaces and their usages. Theoretically, all the three basic modules can be optimized using any kind of method including deep RL. In the implementation, both lane-selection and obstacle-detection modules are rule-based and reasonably optimized for the sake of simplicity. The lane-tracking module is optimized using model-free deep RL. The benefit of such hierarchical implementation over end-to-end training is that basic attribute modules are much easier to optimize compared to end-to-end training of complicated high-level driving policies. 
\begin{table}[b]
\caption{Definition of Driving Tasks}
\label{driving_env_table}
\begin{center}
\begin{tabular}{| c | c | c |}
\hline
Task & Observation & Evaluation \\ \hline
LK & $o_{vh}, o_{ref,i^*}$ & $\Sigma \gamma^t r_{tra},r_{dev}$ \\ \hline
LC & $o_{vh}, \{o_{ref,i}\}, i^*$ & $\Sigma \gamma^t r_{tra},r_{dev}$ \\ \hline
OA & $o_{vh}, \{o_{ref,i}\}, o_{srd}$ & $\Sigma \gamma^t r_{tra},r_{dev},r_{col}$ \\
\hline
\end{tabular}
\end{center}
\end{table}
\begin{table}[b]
\centering
\caption{Definition of RL modules}
\label{rl_module_table}
\begin{center}
\begin{tabular}{| c | c | c |}
\hline
deep RL module & Input & Output \\ \hline
Lane tracking & $o_{vh}$, $o_{ref,i^*}$ & $a_{vh}$\\ \hline
Lane selection & $\{o_{ref,i}\}$, $i^*$ & $o_{ref,i^*}$ \\ \hline
Obstacle detection & $o_{vh}$, $o_{srd}$ & $i^*$ \\
\hline
\end{tabular}
\end{center}
\end{table}

\subsection{DOB-based Tracking Controller}
We design the reference trajectory as a trajectory of $(X, Y,\psi_v,v_x)$. While the speed profile is followed directly by executing corresponding $a_x$, the path defined by the series of $(X, Y, \psi_v)$ is tracked by the tracking controller. To design the controller, first we need to obtain an approximate linear model for the tracking problem. We adopt the constant speed linear bicycle model for lateral dynamics \cite{rajamani2011vehicle}:
\begin{equation}
\dot{x} = A \cdot x + B \cdot \delta, \label{eqn:lbm}
\end{equation}
where the state variable is $x = [v_y\ \psi\ \dot\psi]^T$. Under small angle assumption, $\beta\approx{v_y}/{v_x}$, we approximate $\beta_s$ as: 
\begin{align}
\beta_s\approx\frac{v_x\beta+d_s\dot{\psi}}{v_{x}}=\frac{1}{v_x}v_y+\frac{d_s}{v_x}\dot{\psi}.
\end{align}
Thus:
\begin{align}
\psi_{v_s} = \psi+\beta_s = \left[\frac{1}{v_x}\ 1\ \frac{d_s}{v_x}\right] \left[ {\begin{array}{*{20}{c}}
{{v_y}}\\
\psi \\
{\dot \psi } \end{array}} \right] = C \cdot x. \label{eqn:psi_vs}
\end{align}
Following \cite{rathgeber2014lateral}, the derivative of $\psi_{s,ref}$ is approximated as $\dot{\psi}_{s,ref}\approx{v_x}{\kappa_{s,ref}}$, where $\kappa_{s,ref}$ refers to the curvature of the curve at the reference point. Consequently:
\begin{align}
\Delta{\dot{\psi}_s}&={\dot{\psi}_{v_s}-\dot{\psi}_{s,ref}}\approx \dot{\psi}_{v_s}-v_x\kappa_{s,ref}.
\end{align}
Finally, $\Delta{\dot{y}}_s$ can be approximated as $\Delta{\dot{y}}_s\approx{v_x}{\Delta{\psi_s}}$. 

\begin{figure}[t]
\centering
\includegraphics[scale=0.33]{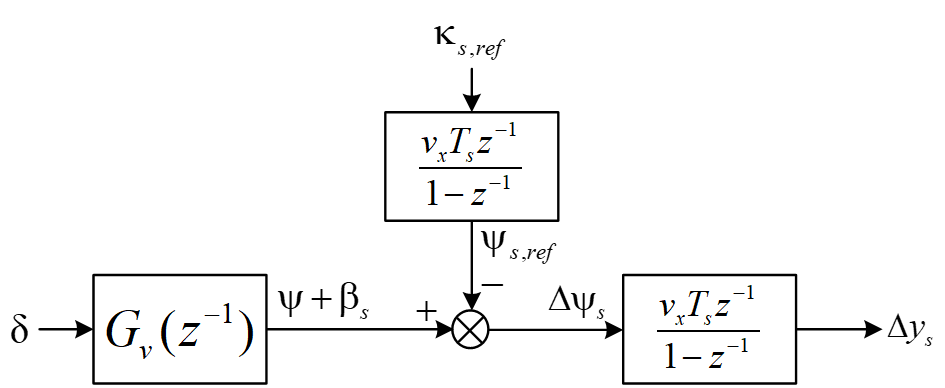}
\caption{Block diagram of linear tracking model}
\label{fig:block_dyn}
\end{figure}

Using the equations above and applying forward Euler discretization, the overall tracking model can be obtained as the block diagram shown in Fig. \ref{fig:block_dyn}, where $T_s$ is the sampling time, $\frac{T_sz^{-1}}{1-z^{-1}}$ is the transfer function of the discretized integrator. $G_v(z^{-1})$ is the vehicle dynamics that maps $\delta$ to $\psi_{v_s}$, of which we can derive the nominal transfer function $G_{nv}(z^{-1})$ using (\ref{eqn:lbm}) and (\ref{eqn:psi_vs}). As shown in the diagram, $\psi_{s,ref}$ can be considered as disturbance to the system.

The robust controller is designed based on the nominal model based on the source vehicle. First, we design a proportional feedback controller as: $u_c = -k_1\Delta{\psi_s}-k_2\Delta{y_s}$. For analysis, we can further write the control law as:
\begin{align}
u_c=-k_2(\frac{k_1}{k_2}+\frac{v_x T_sz^{-1}}{1-z^{-1}})\Delta{\psi_s}=-k_2C_1(z^{-1})\Delta{\psi_s}
\end{align}
where $C_1(z^{-1})=\frac{k_1}{k_2}+\frac{v_xT_sz^{-1}}{1-z^{-1}}$. In the implementation, we hold constant $\frac{k_1}{k_2}$, and tune $k_2$ to achieve stability of the closed-loop system using root locus. We inspect stability for various $v_x$ such that the resulting controller can stabilize the vehicle for the range of velocity in the given driving tasks.

\begin{figure}[t]
\centering
\includegraphics[scale=0.6]{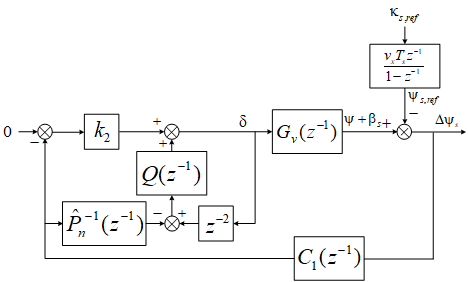}
\caption{Block diagram of closed-loop system}
\label{fig:block_clp}
\end{figure}

Then a DOB is added to the nominal feedback controller. The block diagram for the overall closed-loop system is shown in Fig. \ref{fig:block_clp}. DOB is inserted between $C_1(z^{-1})$ and $k_2$ so that disturbance due to modeling error between $\Delta{y_s}$ and $\Delta{\psi_s}$ can be rejected. $P_n(z^{-1})$ and $\hat{P}_n(z^{-1})$ are the nominal plant and nominal plant without delay, defined as:
\begin{align}
P_n(z^{-1})=z^{-2}\hat{P}_n(z^{-1}) = G_{nv}(z^{-1})C_1(z^{-1})
\end{align}
In our case, the nominal plant $P_n(z^{-1})$ has two timesteps' delay. We make $\hat{P}_n(z^{-1})$ free of delay so that $\hat{P}_n^{-1}(z^{-1})$ is realizable. The closed-loop sensitivity function with DOB is:
\begin{align}
S = \frac{1}{1+k_{2}G_vC_1+z^{-2}(\frac{G_v}{G_{nv}}-1)Q}(1-z^{-2}Q).
\end{align}
We design $Q$ as a second-order low-pass filter to reject low-frequency disturbances, because $\kappa_{s,ref}$ should have relatively low frequency components for smooth reference trajectory. 

When modeling errors exist, we can use the robust control theorem to ensure robust stability \cite{chen2013control}. However, modeling uncertainty between the linear model and the nonlinear model nonlinear model is involved in our problem, which complicates the analysis. Future efforts will be made to approximate the uncertainty bound for robustness specification. 

One last note is that $P$ is time-varying as $v_x$ is not constant in general. In our implementation, we designed the DOB based on the final steady speed of the vehicle in simulations for simplification. In practice, an adaptive DOB can be designed.

\section{Simulation Results}
\subsection{Training of Hierarchical RL Model}

In the off-line training phase, the nominal vehicle dynamics is applied, and the environment includes parallel sinusoidal lanes with lane width of 3 meters. Both the policy network and the value network are three layer fully-connected neural networks. We used the proximal policy optimization (PPO) \cite{schulman2017proximal} algorithm as the model-free deep RL algorithm. Training techniques such as advantage normalization and reparameterization were applied.

In the training phase, the agent achieves successful driving for 1000 timesteps in around 7000 iterations, and the policy optimization converges in around 12000 iterations. After the RL modules are optimized, they can be flexibly assembled to complete different driving tasks, using structures shown in Fig. \ref{fig:hierarchical}. Each deep RL policy is then cascaded with the source environment to get the transferable trajectory planner. The performances of the hierarchical RL policies for all the three tasks are satisfying, which gives us confidence of the transfered deep RL-based high-level trajectory planner. The results will be presented later in Subsection IV-C, for direct comparison with the transfer system.

\subsection{Analysis of DOB-based Tracking Controller}
\begin{figure}[t]
\centering\includegraphics[scale=0.55]{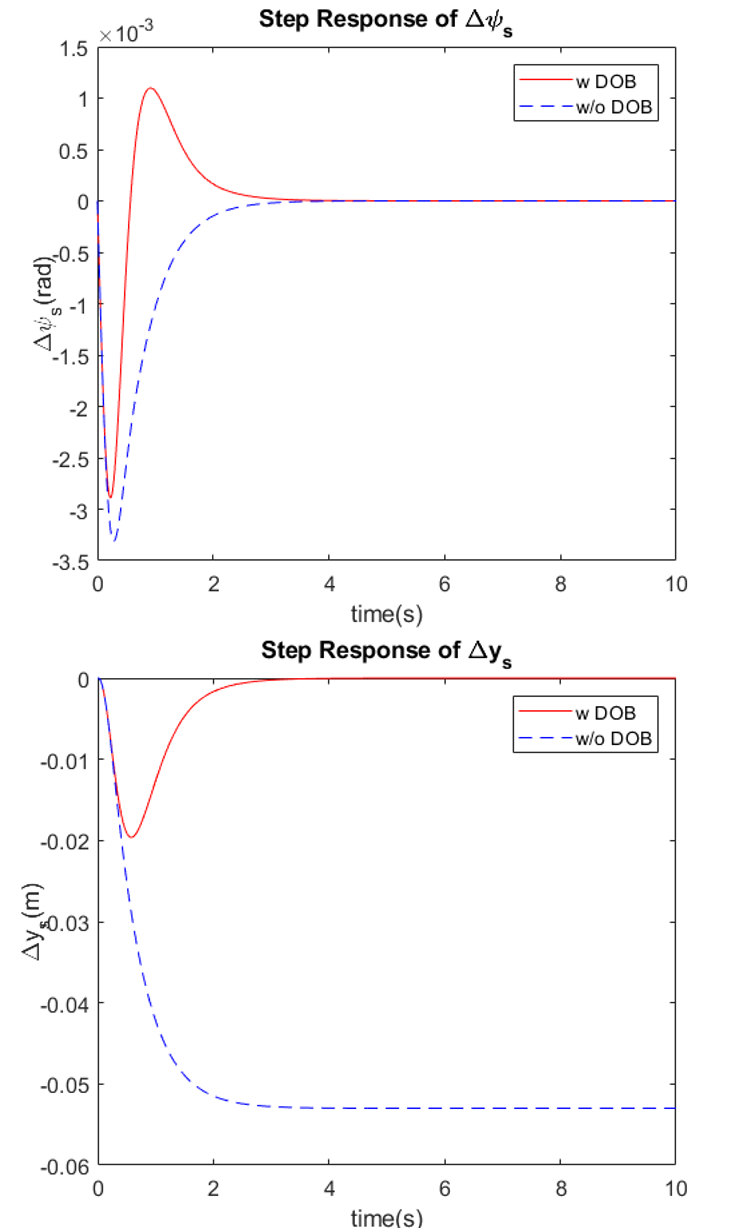}
\caption{Step responses of $\Delta{y_s}$ and $\Delta{\psi}_s$ given a step input of previewed reference curvature $\kappa_{s,ref}$ with a magnitude of $10^{-3} m^{-1}$}
\label{fig:step}
\end{figure}

Given the parameters of the nominal vehicle model, we specified the look-ahead distance as $d_s=15{\rm m}$ and designed the controller. We analyzed the closed-loop system at $v_x=20{\rm m/s}$, which is approximately the final steady speed in our simulations. Fig. \ref{fig:step} shows the step responses of $\Delta{y_s}$ and $\Delta{\psi}_s$ given a step input of previewed reference curvature $\kappa_{s,ref}$ with a magnitude of $10^{-3} m^{-1}$. DOB enables smaller tracking errors for both $\Delta{y}_{s}$ and $\Delta{\psi}_{s}$. Particularly, steady state error of $\Delta{y}_{s}$ is eliminated by adding DOB. 

\subsection{Performance of Overall System}

In this subsection, we evaluated the performance of the proposed RL-RC transfer architecture, and compared it with the baseline RL policies. We tested the models in all the three tasks (i.e., LK, LC, and OA). For source domain tests, we ran the baseline policies and their corresponding RL-RC systems on the nominal vehicle settings with random initial conditions for 10 times. For target domain tests, both driving strategies were deployed and tested from random initial conditions in 10 target settings with modeling gap.

We considered two types of modeling gap: (i) Variation in model parameters. We randomly generated test environments by adding zero-mean and uniformly distributed errors to the parameters of the vehicle model, including vehicle geometry, mass, rotational inertia, tire model parameters, and friction factors. We evaluated the performance on 10 different target vehicles. (ii) External side-force caused by the side slope of the road or wind. We added a constant external force along $Y$ axis in the target environments. We conducted 10 tests for each given magnitude of the external force. We recorded the total discounted rewards and episodic lengths for each test episode, and calculated their means and standard deviations for each testing category. 

\begin{figure}[t]
\centering\includegraphics[scale=0.48]{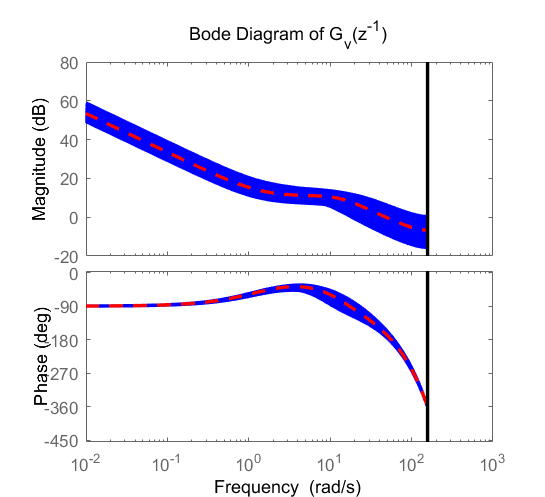}
\caption{Bode plots of $G_v(z^{-1})$ with parameter variation bounded by $20\%$. The blue curves correspond to 100 samples of $G_v(z^{-1})$ with uniformly distributed parameter variation bounded by $20\%$. The red dash curve corresponds to the nominal model.}
\label{fig:bode_uncertain}
\end{figure}

\begin{figure*}[t]
\centering
\includegraphics[scale=0.42]{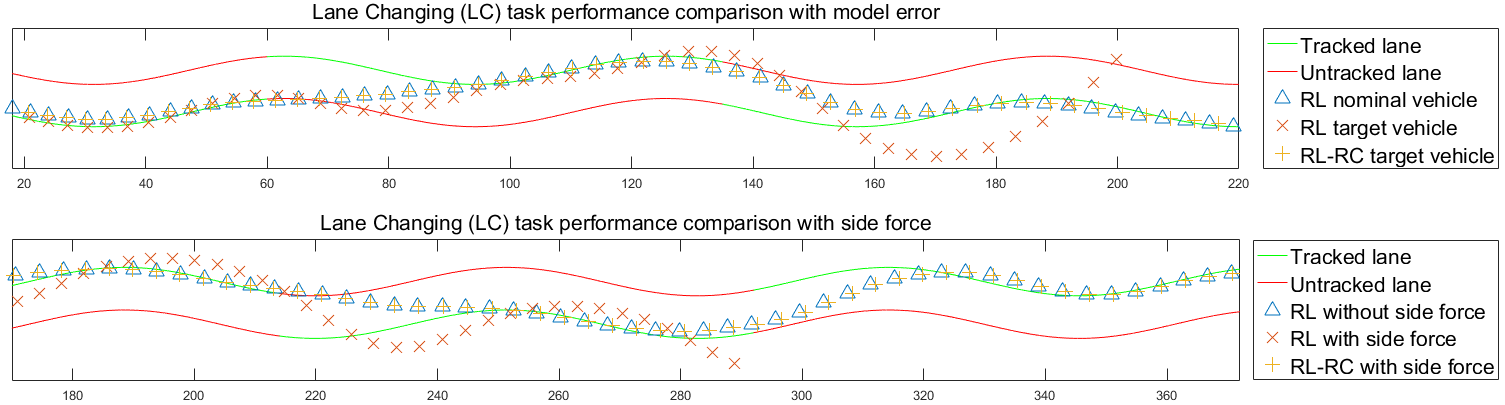}
\caption{Comparison of the driving behaviors for the RL and the RL-RC with modeling gap in the Lane Changing (LC) task}
\label{fig:example}
\end{figure*}

Table. \ref{comparison} gives the overall performance comparison of the baseline RL policy and RL-RC method. Fig. \ref{fig:example} shows two examples illustrating the driving behaviors of the two methods. The target settings have parameter variation bounded by $20\%$ of the nominal values or side force of a magnitude of $5000{\rm N}$. A demonstration video under the same target settings is also provided online \footnote{\url{https://berkeley.box.com/v/ITSC2018}}. To give an intuition of the effects of such modeling gap, Fig. \ref{fig:bode_uncertain} shows the bode plots of 100 samples of $G_v(z^{-1})$ with uniformly distributed parameter variation bounded by $20\%$. The $5000{\rm N}$ side force contributes a maximum lateral acceleration of $2.78{\rm m/s^2}$ to the nominal vehicle.

Table \ref{comparison} and Fig. \ref{fig:example} show the effectiveness of the RL-RC in three aspects: (i) The robustly satisfying performance of the deep RL policies in the source setting is validated, which gives us confidence of the transfered trajectory planner to generate reasonable reference for the tracking controller. (ii) The poor performance of the baseline RL policy in the target settings indicates the incapability of the direct policy transfer. When the baseline policy is applied to a different vehicle or when the target vehicle is affected by a side force, the baseline policy often fails and the target vehicle eventually loses control. (iii) The consistent performance of the RL-RC in both the source and the target domain validates the proposed approach can zero-shoot the transfer tasks. This is thanks to the robust and satisfying tracking performance of the DOB-based tracking controller.

\begin{figure}[t]
\centering\includegraphics[scale=0.25]{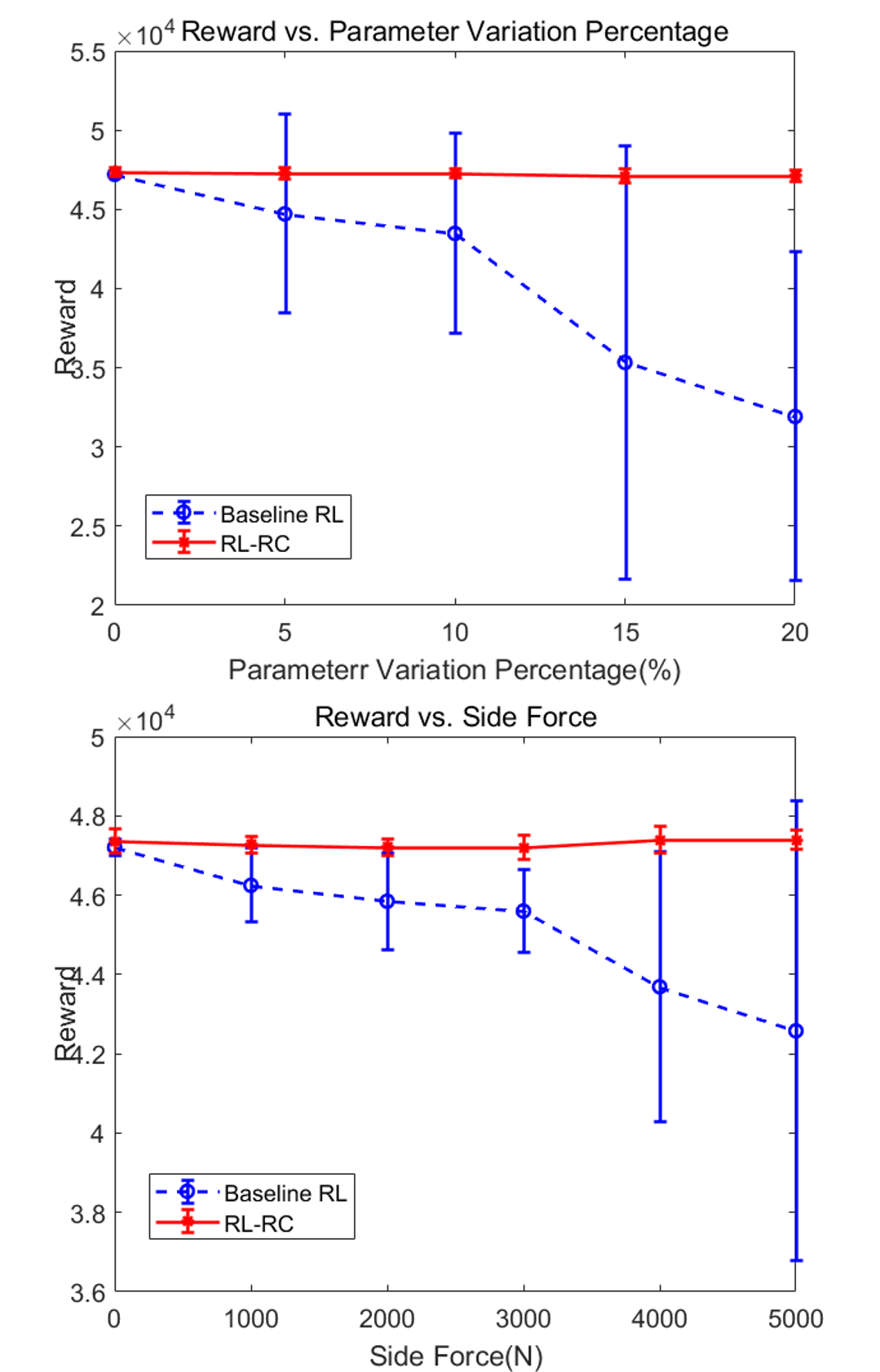}
\caption{Performance (episodic return) of baseline RL policy and RL-RC architecture under increasing modeling gap in Lane Changing (LC) task}
\label{fig:performance}
\end{figure}

Fig. \ref{fig:performance} further compares the performance of the baseline RL policy and RL-RC system under modeling gap of different magnitudes. The curves show the mean value of the reward, and the error-bar indicates the interval with a maximum deviation of one standard deviation from the mean value. The proposed RL-RC framework obtains consistent reward with parameter variation up to $20\%$ or side force of magnitude up to $5000{\rm N}$, whereas the performance of baseline RL policy keeps getting worsen as the modeling gap increases. Decrease in reward can be observed for all tested magnitude of disturbances.

\begin{table*}[b]
\caption{Comparison of the performances of baseline RL policies and RL-RC architecture}
\label{comparison}
\begin{threeparttable}
\centering
\renewcommand\TPTminimum{\linewidth}
\makebox[\linewidth]{\scriptsize%
\begin{tabular}{|c|c|c|c|c|c|c|c|}
\hline
\multirow{2}{*}{Task}&
\multicolumn{3}{c|}{Baseline RL policy}  & \multicolumn{3}{c|}{RL-RC architecture}\cr\cline{2-7}
 & Source & Target - model errors & Target - side force & Source &  Target - model errors &  Target - side force \\ \hline
\multirow{2}{0.2cm}{LK}&
 $1000\pm{0}$ & $926.3\pm{177.8}$ & $1000\pm{0}$ & $1000\pm{0}$ & $1000\pm{0}$ & $1000\pm{0}$ \cr
& $49418.0\pm{270.7}$ & $44644.5\pm{9538.6}$ & $47102.9\pm{542.2} $ & $48688.1\pm{221.3}$& $48956.0\pm{362.8}$ & $48959.2\pm{110.4}$\\ \hline
\multirow{2}{0.2cm}{LC}&
 $1000\pm{0}$ & $775.8\pm{239.3}$ & $963.0\pm{111.0}$ & $1000\pm{0}$ & $ 1000\pm{0}$ & $1000\pm{0}$\cr
& $47209.7\pm{213.3}$ & $33990.2\pm{12007.3}$ & $42578.0\pm{5800.9}$ & $47363.2\pm{303.2}$ & $47150.9\pm{361.1}$ & $47385.2\pm{241.4}$\\ \hline
\multirow{2}{0.2cm}{OA}&
$1000\pm{0}$ & $749.0\pm{239.1}$ & $735.3\pm{256.5}$ & $1000\pm{0}$ & $1000\pm{0}$ & $1000\pm{0}$\cr
& $47348.1\pm{318.4}$ & $33283.7\pm{11846.1}$ & $31655.7\pm{12017.1}$ & $47661.8\pm{214.7}$ & $47521.2\pm{319.1}$ & $47694.8\pm{374.8}$\\ \hline
\end{tabular}}
\begin{tablenotes}
\item[1] LK stands for lane keeping. LC stands for lane changing. OA stands for obstacle avoidance.
\item[2] Data is presented in form of ${\rm mean}\pm{\rm std}$. In each cell, episodic length and total reward are listed from top to bottom.
\end{tablenotes}
\end{threeparttable}
\end{table*}

\section{Conclusion}
We proposed a generic framework for deep RL driving policy transfer using RC techniques. The deep RL policy trained in the source domain is used to generate a finite-horizon previewed reference trajectory of interpretable kinematic features. In the target domain, a robust controller is designed to track the reference trajectory so that disturbances induced by the modeling gap can be rejected. We implemented the framework for three driving tasks in term of a hierarchical RL policy and a DOB-based robust tracking controller. Simulation results showed that the proposed RL-RC architecture can achieve zero-shot consistent performance in the target domain with certain level of parameter variation or external side force. Meanwhile, performance of the baseline RL policy is devastated by the modeling gap. Future efforts will be made to investigate the uncertainty bound introduced by model linearization so that robust stability of the DOB-based controller can be guaranteed. Moreover, the proposed method will be implemented to achieve policy transfer from simulated environment to a real vehicle. Also the current method only considers the variation of the system dynamics. Furthermore, other kinds of modeling gap can affect the policy transfer performances (e.g. number of obstacles and behavior of surrounding vehicles). We will extend the current method to solve more general policy transfer problem for autonomous driving tasks. 
\addtolength{\textheight}{-0cm}   
\bibliography{reference}
\bibliographystyle{ieeetr}

\addtolength{\textheight}{+0cm}
\newpage

\end{document}